\documentclass[letterpaper]{article} % DO NOT CHANGE THIS
\usepackage{aaai25}  % DO NOT CHANGE THIS
\usepackage{times}  % DO NOT CHANGE THIS
\usepackage{helvet}  % DO NOT CHANGE THIS
\usepackage{courier}  % DO NOT CHANGE THIS
\usepackage[hyphens]{url}  % DO NOT CHANGE THIS
\usepackage{graphicx} % DO NOT CHANGE THIS
\usepackage{natbib}  % DO NOT CHANGE THIS AND DO NOT ADD ANY OPTIONS TO IT
\usepackage{caption} % DO NOT CHANGE THIS AND DO NOT ADD ANY OPTIONS TO IT
\usepackage{algorithm}
\usepackage{algorithmic}
\usepackage{enumitem}

\usepackage{newfloat}
\usepackage{listings}
\DeclareCaptionStyle{ruled}{labelfont=normalfont,labelsep=colon,strut=off} % DO NOT CHANGE THIS
\floatstyle{ruled}
\newfloat{listing}{tb}{lst}{}
\floatname{listing}{Listing}
\usepackage[T1]{fontenc}
\usepackage{enumitem}
\usepackage{booktabs}
\usepackage{multirow}
\usepackage{comment}
\usepackage[
disable
]{endfloat}

\title{Retrieving Versus Understanding Extractive Evidence in Few-Shot Learning}
\author {
    Karl Elbakian\textsuperscript{\rm 1},
    Samuel Carton\textsuperscript{\rm 1}
}
\affiliations {
    \textsuperscript{\rm 1}University of New Hampshire\\
    karl.elbakian@unh.edu, samuel.carton@unh.edu
}

\begin{document}
\maketitle
\begin{abstract}
% We analyze the effect of extractive rationalization on large language model prediction in a few-shot setting. Specifically, we measure the extent to which model prediction errors are associated with rationalization errors with respect to gold-standard human-annotated extractive rationales for five datasets. We perform two ablation studies to investigate when both label prediction and evidence retrieval errors can be attributed to qualities of the relevant evidence. We find that there is a strong empirical relationship between model prediction and model rationalization, and that while it is difficult to modify model rationalization via exemplars, it is easier to manipulate model interpretation of rationalized evidence in this manner, leading to contradictory results in some cases.
A key aspect of alignment is the proper use of within-document evidence to construct document-level decisions. We analyze the relationship between the retrieval and interpretation of within-document evidence for large language models in a few-shot setting. Specifically, we measure the extent to which model prediction errors are associated with evidence retrieval errors with respect to gold-standard human-annotated extractive evidence for five datasets, using two popular closed proprietary models. We perform two ablation studies to investigate when both label prediction and evidence retrieval errors can be attributed to qualities of the relevant evidence. We find that there is a strong empirical relationship between model prediction and evidence retrieval error, but that evidence retrieval error is mostly not associated with evidence interpretation error--a hopeful sign for downstream applications built on this mechanism.
\end{abstract}

\begin{links}
    \link{Code}{https://github.com/kelbakian/llm-rationale-fidelity}
    % \link{Datasets}{https://aaai.org/example/datasets}
    % \link{Extended version}{https://aaai.org/example/extended-version}
\end{links}

\section{Introduction}

AI alignment refers to the goal of ensuring that model output is aligned with human intents and values \cite{shen_large_2023,anwar_foundational_2024,shen_towards_2024}. One key element of alignment is \textbf{verification}, the ability to confirm that a model's predictions have indeed accorded with those intents and values. 
% AI interpretability plays a key role in pursuing this goal by offering methods to audit the reasoning underlying model predictions, and inspect them at these lower levels for signs of misalignment. 
% For better or worse, one of the major application areas for large language models (LMs) is in saving human labor by automating or assisting decision-making over input documents,
When we use models to automate or assist human-affecting tasks such as moderation \cite{kumar_watch_2024}, resume screening \cite{gan_application_2024}, grading \cite{pinto_large_2023}, or medical decision-making \cite{thirunavukarasu_large_2023}, we want a human auditor to be able to review their decisions for mistakes or pathologies of behavior such as bias or the use of spurious evidence. 
% we want them to result from at least the facsimile of a reasoned process that can be inspected, verified, and, if necessary, contested by a human auditor.
% Interpretability offers tool for performing this auditing. 
Verification has traditionally been one of the major goals of model interpretability \cite{fok_search_2023}. Implicitly, the assumption underlying this function is that it is easier for a human auditor to catch model mistakes at the explanation level and propagate them upward to an appropriate skepticism about the model's overall prediction, than to inspect that prediction alone.

With the rise of large language models (LMs), the discourse on AI interpretability has turned towards methods that take advantage of their emergent capabilities. Free-text explanations can provide clear, comprehensible, and interactive descriptions of why an LM made a certain prediction \cite{singh_rethinking_2024}, while work in LM reasoning such as Chain-of-Thought \cite{wei_chain_2022} and its descendants force the model to break its reasoning into discrete steps which can be individually inspected. 
% Mechanistic interpretability seeks to identify latent algorithmic ``circuits'' encoded in the parameters of the model \cite{olah_feature_2017,wang_interpretability_2022}. 
What unites these approaches is that they are \textbf{abstractive}, synthesizing the raw input into a concise and human-comprehensible form. 

By contrast, \textbf{extractive} approaches to interpretability such as rationale models \cite{lei_rationalizing_2016} or LIME \cite{ribeiro_why_2016} have fallen somewhat out of vogue because of their inexpressiveness compared to abstractive approaches \cite{siegel_probabilities_2024,hu_think_2023}, as well as the parametric and/or computational overhead they add to models which are already large, unwieldy, and in some cases accessible via API only. However, some recent work has examined the ability of LMs to generatively mimic extractive explanations, or ``self-rationalize'', via prompting, finding them of comparable quality to traditional methods \cite{huang_can_2023}. 

\begin{figure}[h]
% \vspace{-10pt}
    \centering
    \includegraphics[width=0.45\textwidth]{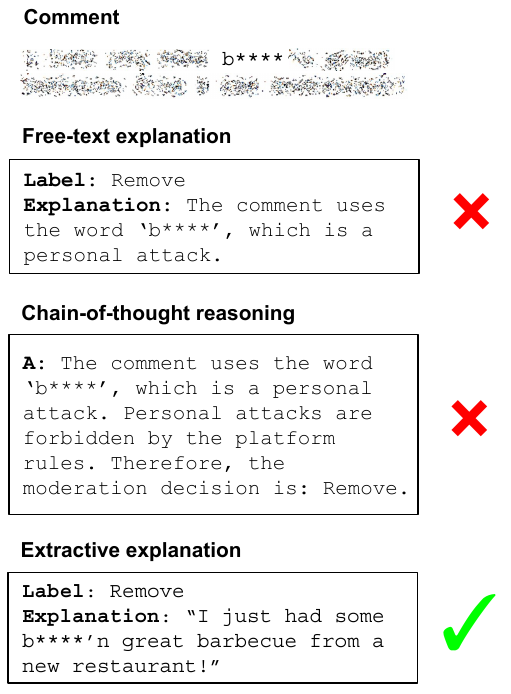}
    \caption{Artificial examples of abstractive and extractive explanations for an erroneous moderation prediction. Only the extractive explanation provides a basis for refuting it. }
    \label{fig:explanation_example}
\end{figure}

However, when we want to verify an LM's prediction over an input document, there are basic questions that are more appropriately answered by extractive approaches than abstractive ones. Namely: \textbf{what evidence from the input document is the model using as the basis for its decision, and does that evidence support its predicted label?} Fully abstractive explanations don't directly elucidate the relationship between the within-text evidence and label, and can in fact obfuscate it. Figure \ref{fig:explanation_example} shows a simple example where abstractive explanations of a moderation decision are inadequate for refuting it. 
% To verify or refute a model prediction over an input document, we must know what evidence within the document the model used, and judge whether that evidence justifies that prediction. 
% No matter how fluent and persuasive a free-text explanation the model can construct, its prediction must ultimately be grounded in the content of the input document, and if we want to verify that prediction we must be able to assess whether it is supported by that content. A free-text explanation may help clarify or contextualize that evidence-prediction relationship, but it may also misrepresent it, as it is just as liable to hallucination as any other LM output, and may not be faithful to the original prediction \cite{lyu_towards_2024}. 

The problem of verification is related to the problem of faithfulness \cite{jacovi_towards_2020}, the idea that a model's explanation should be coupled with, and thus display the true underlying logic of, its prediction. Prompt-based explanations, abstractive or extractive, are intrinsically unfaithful by this definition, as they are just a proximal generation by the model alongside the label \cite{lyu_faithful_2023, turpin_language_2023}. For the purpose of verification, however, we are less interested in answering ``what does the explanation tell us about the model's prediction?'' than ``what does the correctness of the model's explanation tell us about the correctness of its prediction?'' Analogous to the concept of internal vs. external validity in experimental design \cite{mcdermott_internal_2011}, these two questions can be viewed as \textbf{internal vs. external faithfulness}, respectively.

In this paper we examine the external faithfulness of prompt-based self-rationalization in two prominent large proprietrary language models: GPT-4 \cite{openai_gpt-4_2023} and Gemini \cite{team_gemini_2023}. Our high-level research question is: \textbf{Can large language models reliably and meaningfully identify within-document evidence for their predictions?} With an eye towards developing downstream human-model interaction systems based on this mechanism, we investigate the following specific questions: 
(1) Can LMs reliably quote evidence snippets from an input document without mistakes?;
(2) Does forcing self-rationalization impact model accuracy?;
(3) Does operation order (explain-then-predict versus predict-then-explain) affect performance?;
(4) Are label prediction errors correlated with evidence retrieval errors, relative to human gold-standard evidence?;
(5) What types of evidence retrieval errors cause label prediction errors?;
(6) Under what circumstances do LMs fail to retrieve evidence?
These last two questions are of key importance for the design of any human-model collaborative system based on extractive evidence as an underlying mechanism. If the characteristic failure mode for the model is to identify the correct pertinent evidence and then simply to misinterpret it, this is much more correctable, via either prompt engineering approaches such as self-consistency \cite{wang_self-consistency_2023} or human inspection, than if errors stem from missing key evidence entirely. We focus exclusively on \textbf{extractive} self-rationalization for the reasons outlined above: they are easier to assess for correctness (and thus faithfulness), and invite fewer pitfalls as a verification mechanism.

We experiment with five datasets for which gold standard human-annotated extractive evidence is available, covering a wide range of tasks: MultiRC \cite{khashabi_looking_2018}, SciFact \cite{wadden_scifact-open_2022}, WikiAttack \cite{carton_extractive_2018}, Evidence Inference \cite{deyoung_evidence_2020}, and HealthFC \cite{vladika_healthfc_2024}. For a sample of each dataset, we run experiments prompting the model to self-rationalize label predictions under varying conditions, comparing and contrasting the result to the gold-standard evidence in each dataset. We find broadly that (1) these  models \textbf{can} reliably quote within-text evidence; (2) self-rationalization mostly does \textbf{not} effect label accuracy,  (3) operation order does \textbf{not} matter; (4) label prediction error \textbf{is} highly correlated with evidence error for most datasets; (5) label error is more commonly linked with capturing confounding evidence rather than missing key evidence; and finally (6) missing key evidence is mostly commonly linked to the presence of redundant evidence rather than more intractable interpretation issues. All of these are positive outcomes, suggesting the potential for LM extractive self-rationalization as a mechanism for powering downstream applications. All code and experimental results can be found in our github repository.

\section{Related Work}

% \paragraph{Interpretability in large language models.}

Recent interpretability work has tended to focus on abstractive approaches such as posthoc free-text explanations \cite{singh_rethinking_2024, zhu_explanation_2024}, or explanations as a byproduct of explicit reasoning processes like Chain-of-Thought \cite{wei_chain_2022, lanham_measuring_2023}. However, one of the major goals of interpretability is verification \cite{fok_search_2023}, the implicit assumption being that it is easier to recognize an erroneous explanation than an erroneous label. This mode requires explanations to be faithful \cite{jacovi_towards_2020} to the overall prediction, but this quality is difficult to measure in abstractive approaches \cite{agarwal_faithfulness_2024, siegel_probabilities_2024}.

Extractive approaches are less problematic in this regard because they directly attribute the prediction to evidence within the input. Even if this evidence is not technically faithful to the model's prediction, an observer can still assess whether it truly supports the label without being potentially beguiled by misleading abstractive generations of the model. Traditional approaches to identifying extractive evidence, such as the rationale model architecture \cite{lei_rationalizing_2016} or the LIME perturbation method \cite{ribeiro_why_2016}, add impractical levels of computational overhead to already large models, so recent work has investigated whether LMs can be prompted to produce such attributions as a generative output \cite{huang_can_2023,hu_think_2023, majumder_knowledge-grounded_2022}. An especially relevant recent work is \cite{madsen_are_2024}, which investigates the internal faithfulness of several types of extractive explanations produced by three open LMs. Our approach follows this work in directly prompting LMs to produce extractive evidence for their predictions. To assess whether LMs can identify the ``correct'' evidence, we use datasets with gold-standard evidence annotations. The ERASER collection \cite{deyoung_eraser_2019} includes a number of these datasets, and \cite{wiegreffe_teach_2021} surveys yet more. 

Finally, our goal of retrieving relevant evidence buried in potentially long documents is similar to that of``needle-in-the-haystack'' evaluations \cite{dhinakaran_needle_2024}, which ask questions about evidence manually inserted into a long context. Where we differ is in applying this approach to naturally-occurring evidence that the model may not be able to properly interpret, rather than artificially-inserted evidence the model is assumed to be able to comprehend if it can find it.

\section{Datasets}

We analyze five datasets for which gold-standard human-annotated extractive evidence is included alongside ground truth labels: MultiRC \cite{khashabi_looking_2018}, SciFact \cite{wadden_scifact-open_2022}, and WikiAttack \cite{carton_extractive_2018}, Evidence Inference \cite{deyoung_evidence_2020}, and HealthFC \cite{vladika_healthfc_2024}. To reduce API access costs, we perform our analysis on randomly-sampled 300-item subsets of the development set for each dataset.
% except WikiAttack.

% We evaluate the effects of various few-shot exemplar constructions on two ERASER benchmark datasets - SciFact and MultiRC, as well as the WikiAttack dataset.
 
\paragraph{SciFact (SF)}
\cite{wadden_scifact-open_2022} is a scientific claim verification dataset, involving identifying whether abstracts from the research literature either support or refute a given scientific claim. The dataset contains 1,400 expert-written claims, paired with evidence-containing abstracts annotated with veracity labels and sentence-level rationales. Rationales are direct spans of text from the document which support the class label given. Each claim has a class label of \textit{SUPPORT}, \textit{NO INFO}, or \textit{CONTRADICT}. 

\paragraph{MultiRC (MRC)}
\cite{khashabi_looking_2018} is a reading comprehension dataset containing sets of short paragraphs and questions that depend on information from multiple sentences within the paragraph. Each paragraph-question pair contains five answers, with a variable number of correct answer-options. Additionally, answer-options do not have to be a span from the text. The dataset contains sentence-level rationales in the form of relevant sections of the paragraph, as well as \textit{True} or \textit{False} labels for each answer candidate.

\paragraph{WikiAttack (WA)}
\cite{carton_extractive_2018} contains excerpts of Wikipedia comment threads original included in the WikiToxic dataset of \cite{wulczyn_ex_2017} which are scanned for instances of personal attacks or harassment. Class labels are given as either \textit{attack} or \textit{nonattack}, and rationales for comments labeled as attack are given in the form of spans from the comment.

\paragraph{Evidence Inference (EI)}
\cite{deyoung_evidence_2020} dataset contains scientific articles and queries asking to compare the relative effectiveness of two treatments with respect to a given outcome. Labels include \textit{significantly increased}, \textit{significantly decreased}, and \textit{no significant difference}. Annotations are in the form of quotes from the text which support the label.

\paragraph{HealthFC (HFC)}
\cite{vladika_healthfc_2024} is a medical domain QA dataset. Given a health-related claim, verdicts are decided from a systematic review or clinical trial document, as one of \textit{supported}, \textit{refuted}, and \textit{not enough information}. Additionally, annotations containing up to five quotes from the document are provided. After reviewing the dataset, we found discrepancies between the original, expert-annotated, German verdicts, and their English label mappings. Thus, we performed a re-labeling of the dataset, first by prompting GPT-4 to generate an English label given the query and German verdict, and then by manually labeling each item in the dataset, using a holistic review of the original English label, German verdict, cited evidence, and full document text. The adjusted labels are included in our repository.

\begin{figure}[]
% \vspace{-10pt}
    \centering
    \includegraphics[width=0.474\textwidth]{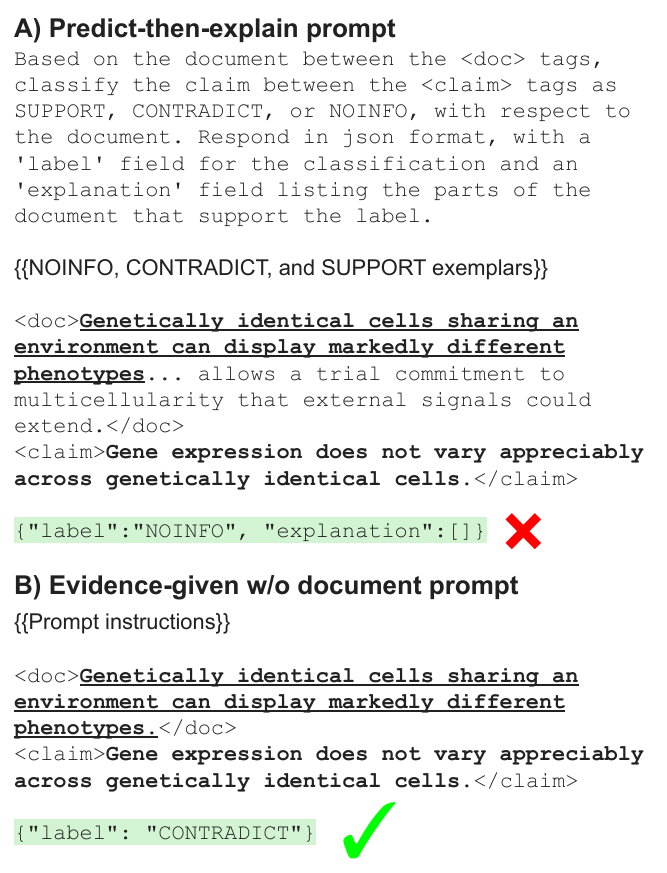}
    \caption{Two examples of GPT-4 prompts on the same SciFact item. Model output highlighted in green. Human-annotated evidence underlined, claim bolded. The model misses the evidence and mislabels the document in the predict-then-explain setting, but correctly labels it when the evidence is provided, even without its surrounding context.}
    \label{fig:scifact_prompt_example}
\end{figure}

\section{Implementation Details}

\paragraph{Model details.} 
Because of their to-date superior few-shot performance compared to open models, we focus on two closed proprietary models: GPT-4 and Gemini. 
For GPT-4, the \textit{gpt-4-0613} version is used for MultiRC, WikiAttack, and SciFact. Because of long input lengths, \textit{gpt-4-turbo} is used for Evidence Inference and HealthFC. All GPT model temperatures are set to the default 0.7. For Gemini, the \textit{gemini 1.5 pro} model version is used for all datasets, with temperature set to the default 1.0 
%ACL cut: For the fine-tuned baseline, we use the RoBERTa-large \cite{liu_roberta_2019} implementation available on Hugging Face, training for 10  epochs with early stopping. 

% \paragraph{Self-Rationalized prediction.} Figure \ref{fig:scifact_prompt_example}A illustrates the prompt used to perform rationalized prediction. We produce predictions and extractive rationales concurrently in an ``explain-and-predict'' manner, introducing the possibility of unfaithful rationales with no actual relationship to the model's label prediction. Part of our analysis examines this relationship empirically. The prompt presents sampled exemplars in JSON format, and requests model output in this format. 

\paragraph{Exemplar sampling.} All few-shot conditions involve sampling two exemplars for each possible class, resulting in 4-shot learning for MultiRC and WikiAttack, and 6-shot learning for SciFact, Evidence Inference, and HealthFC. We randomly sample and shuffle exemplars independently for each item. For MultiRC, SciFact, Evidence Inference, and HealthFC, exemplars are sampled from the training set. For WikiAttack, where human rationales are unavailable for the training set, they are instead sampled from the development set, resulting in effectively 250-fold cross-validation.

\paragraph{Experimental conditions and prompting.} 

We compare model performance across 7 different prompting conditions: (1) \textbf{zero-shot}, with neither exemplars nor self-rationalization; (2) standard \textbf{few-shot} with no rationalization; (3) \textbf{predict then explain}, where the model is asked to first make a label prediction then provide evidence to justify its answer; (4) \textbf{explain then predict}, where the model is asked the reverse order of the former condition; (5) \textbf{evidence given}, in which the human-annotated evidence is provided as part of the input prompt; (6) \textbf{evidence given without document}, where the human annotated evidence is provided without the context of the surrounding document; and (7) \textbf{evidence occluded}, where the human-annotated evidence is removed without replacement from the document. 

Fig. \ref{fig:scifact_prompt_example} shows example prompts for the ``predict-then-explain'' and ``evidence given without document'' conditions. Model output is solicited in \texttt{JSON} format using \texttt{<claim>} and \texttt{<doc>} tags to denote different elements of the input.  Prompt scaffolds can be found in our repository.

\section{Results}

% In order to fully explore the effect of imposed rationalization and the relationship between rationalization and prediction performance, we evaluate up-to-date versions of both a closed model (GPT-4) and an open-weight model (Gemini-1.5-pro), 
% Across our analysis, we use several prompting conditions: (1) \textbf{zero-shot}, with neither exemplars nor self-rationalization; (2) standard \textbf{few-shot} with no rationalization; (3) \textbf{predict then explain}, where the model is asked to first make a label prediction then provide evidence to justify its answer; (4) \textbf{explain then predict}, where the model is asked the reverse order of the former condition; (5) \textbf{evidence given}, in which the human-annotated evidence is provided as part of the input prompt; (6) \textbf{evidence given without document}, where the human annotated evidence is provided without the context of the surrounding document; and (7) \textbf{evidence occluded}, where the human-annotated evidence is removed without replacement from the document. 

% We address our six research questions in order:

%\subsubsection{Can language models reliably quote evidence from an input document?}
\begin{table}[t]
\centering
\begin{tabular}{ccc}
\hline
\multirow{2}{*}{\textbf{Dataset}} & \multicolumn{2}{c}{\textbf{Failure Rate (\%)}} \\ \cline{2-3} 
                                  & GPT-4     & Gemini-1.5      \\ \hline
MRC                               & 2.00              & 2.33                    \\ 
SF                                & 0.33               & 0.00                       \\ 
WA                                & 0.00               & 4.33                    \\ 
EI                                & 40.33              & 25.0                   \\ 
HFC                               & 15.33               & 0.33                    \\ \hline
\end{tabular}
\caption{Model evidence quoting failure rates in the explain-then-predict condition}
\label{tab:fail_rate_tab}
\end{table}

%\subsection{Can language models reliably quote evidence from an input document?}

\begin{table}[h]
\begin{tabular}{cccc}
\hline
\multirow{2}{*}{\textbf{Dataset}} & \multirow{2}{*}{\textbf{Condition}}                                      & \multicolumn{2}{c}{\textbf{Label accuracy}} \\ \cline{3-4} 
                                  &                                                                          & \textbf{GPT-4}  & \textbf{Gemini-1.5}  \\ \hline
MRC                               & Zero-shot                                                                & 0.887           & 0.840                \\
                                  & Few-shot                                                                 & 0.893           & 0.892                \\
                                  & Predict then explain                                                     & 0.897           & 0.885                \\
                                  & Explain then predict                                                     & 0.861           & 0.874                \\
                                  & Evidence given                                                        & 0.887           & 0.885                \\
                                  & \begin{tabular}[c]{@{}c@{}}Evidence given\\ w/o document\end{tabular} & 0.827           & 0.797                \\
                                  & Evidence occluded                                                        & 0.663           & 0.666                \\ \hline
\multirow{7}{*}{SF}               & Zero-shot                                                                & 0.853           & 0.850                \\
                                  & Few-shot                                                                 & 0.843           & 0.820                \\
                                  & Predict then explain                                                     & 0.870           & 0.837                \\
                                  & Explain then predict                                                     & 0.833           & 0.841                \\
                                  & Evidence given                                                        & 0.947           & 0.947                \\
                                  & \begin{tabular}[c]{@{}c@{}}Evidence given\\ w/o document\end{tabular} & 0.907           & 0.890                \\
                                  & Evidence occluded                                                        & 0.633           & 0.587                \\ \hline
\multirow{7}{*}{WA}               & Zero-shot                                                                & 0.743           & 0.674                \\
                                  & Few-shot                                                                 & 0.777           & 0.703                \\
                                  & Predict then explain                                                     & 0.790           & 0.728                \\
                                  & Explain then predict                                                     & 0.743           & 0.690                \\
                                  & Evidence given                                                        & 0.813           & 0.832                \\
                                  & \begin{tabular}[c]{@{}c@{}}Evidence given\\ w/o document\end{tabular} & 0.917           & 0.901                \\
                                  & Evidence occluded                                                        & 0.533           & 0.563                \\ \hline
\multirow{7}{*}{EI}               & Zero-shot                                                                & 0.832           & 0.765                \\
                                  & Few-shot                                                                 & 0.809           & 0.767                \\
                                  & Predict then explain                                                     & 0.870           & 0.839                \\
                                  & Explain then predict                                                     & 0.888           & 0.831                \\
                                  & Evidence given                                                        & 0.907           & 0.880                \\
                                  & \begin{tabular}[c]{@{}c@{}}Evidence given\\ w/o document\end{tabular} & 0.920           & 0.893                \\
                                  & Evidence occluded                                                        & 0.721           & 0.698                \\ \hline
\multirow{7}{*}{HFC}              & Zero-shot                                                                & 0.763           & 0.803                \\
                                  & Few-shot                                                                 & 0.733           & 0.787                \\
                                  & Predict then explain                                                     & 0.835           & 0.809                \\
                                  & Explain then predict                                                     & 0.783           & 0.819                \\
                                  & Evidence given                                                        & 0.947           & 0.857                \\
                                  & \begin{tabular}[c]{@{}c@{}}Evidence given\\ w/o document\end{tabular} & 0.850           & 0.863                \\
                                  & Evidence occluded                                                        & 0.677           & 0.687                \\ \hline
\end{tabular}
\caption{Label Accuracy for different prompting paradigms}
\label{tab:label-acc-comparison}
\end{table}

\begin{table*}[t]
\centering
\begin{tabular}{cccccccccc}
\hline
                                                      & \multicolumn{1}{l}{\textbf{}}                       & \multicolumn{4}{c}{\textbf{Correct predictions}}                                 & \multicolumn{4}{c}{\textbf{Incorrect predictions}}                               \\ \hline
\multicolumn{1}{l}{\multirow{2}{*}{\textbf{Dataset}}} & \multicolumn{1}{l}{\multirow{2}{*}{\textbf{Model}}} & \multirow{2}{*}{\textbf{Count}} & \multicolumn{3}{c}{\textbf{Evidence}}         & \multirow{2}{*}{\textbf{Count}} & \multicolumn{3}{c}{\textbf{Evidence}}         \\ \cline{4-6} \cline{8-10} 
\multicolumn{1}{l}{}                                  & \multicolumn{1}{l}{}                                &                                 & \textbf{F1} & \textbf{Recall} & \textbf{Prec.} &                                 & \textbf{F1} & \textbf{Recall} & \textbf{Prec.} \\ \hline
\multirow{2}{*}{MRC}                                  & GPT-4                                               & 253                             & 0.71        & 0.68            & 0.84           & 41                              & 0.59**      & 0.62            & 0.63***        \\
                                                      & Gemini-1.5                                          & 256                             & 0.67        & 0.63            & 0.86           & 37                              & 0.67        & 0.53            & 0.67           \\ \hline
\multirow{2}{*}{SF}                                   & GPT-4                                               & 249                             & 0.76        & 0.77            & 0.84           & 50                              & 0.11***     & 0.75            & 0.12***        \\
                                                      & Gemini-1.5                                          & 252                             & 0.80        & 0.77            & 0.86           & 48                              & 0.14***        & 0.60            & 0.45           \\ \hline
\multirow{2}{*}{WA}                                   & GPT-4                                               & 223                             & 0.85        & 0.94            & 0.85           & 77                              & 0.15***     & 0.92            & 0.18***        \\
                                                      & Gemini-1.5                                          & 198                             & 0.82        & 0.89            & 0.87           & 89                              & 0.16***        & 0.90            & 0.23           \\ \hline
\multirow{2}{*}{EI}                                   & GPT-4                                               & 159                             & 0.62        & 0.62            & 0.69           & 20                              & 0.38***     & 0.33***         & 0.48**         \\
                                                      & Gemini-1.5                                          & 187                             & 0.83        & 0.61            & 0.63           & 38                              & 0.81        & 0.41            & 0.39           \\ \hline
\multirow{2}{*}{HFC}                                  & GPT-4                                               & 199                             & 0.37        & 0.38            & 0.44           & 55                              & 0.29        & 0.29            & 0.36           \\
                                                      & Gemini-1.5                                          & 245                             & 0.53        & 0.38            & 0.48           & 54                              & 0.53        & 0.35            & 0.36           \\ \hline
\end{tabular}
\caption{Evidence rationalization performance for correct vs incorrect label predictions in the \textit{explain then predict} condition. Asterisks denote significant differences between values using a 2-sided t-test; *:\textit{p}\textless0.1; **:\textit{p}\textless0.05; ***\textit{p}\textless0.005}
\label{tab:correctness_table}
\end{table*}

 \begin{figure*}[h]
    \centering
    \includegraphics[width=.8\textwidth]{ 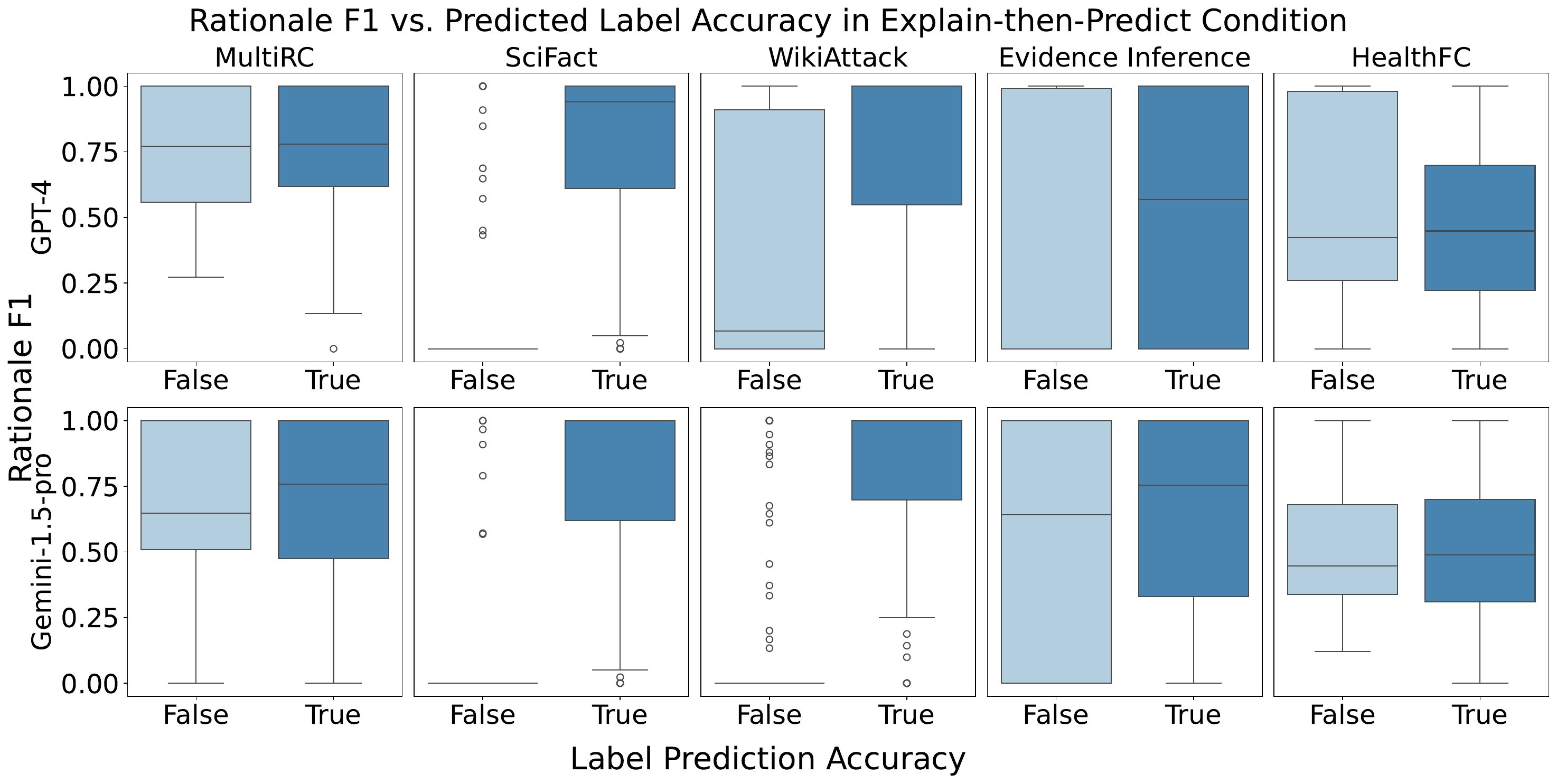}
    \caption{Box-and-whisker plots of label prediction error versus mean predicted evidence rationale F1 for all five datasets in the \textit{explain then predict} condition.}
    \label{fig:label_rationale_whisker_plots}
 \end{figure*}

\subsection{Can language models reliably quote evidence from the input document?}
Table \ref{tab:fail_rate_tab} shows self-rationalization failure rates across the explain-then-predict condition, where the model fails to return any valid quote from an input text, using string substring matching. 
We observe mostly very low error rates, with the two high error rates for GPT-4 are associated with the \texttt{GPT-4 turbo} variant model, which is known to be weaker than the base model. Gemini also shows a high rate of error for Evidence Inference. The results show that, by and large, the strongest contemporary models can reliably quote evidence from the input document. Evidence Inference is the outlier, possibly due to long context lengths.

\subsection{Does self-rationalization impact model accuracy?}
\label{subsec:explaining_errors}
Table \ref{tab:label-acc-comparison} summarizes the overall label prediction accuracy results for all conditions. For two datasets, Evidence Inference and HealthFC, forcing the model to support its prediction has a substantial positive effect (+7.9/5.0 for GPT-4 and +6.4/3.2 for Gemini-1.5, respectively). This effect is marginal for the other three datasets, even sometimes having a negative effect. This effect interestingly is the opposite of the failure rate results, suggesting that longer documents benefit more from this prompting requirement. 
% Next, all datasets except for MultiRC see an increase in performance when provided with the human-annotated evidence for both models. This effect is more pronounced for Gemini-1.5, except in the case of HealthFC. Lastly, we note a uniformly sharp decrease in performance across all datasets when the human-annotated evidence is redacted from the input document.

%\subsubsection{Effect of prompt operation order}
\subsection{Effect of prompt operation order}
Table \ref{tab:label-acc-comparison} also provides a comparison between asking the model to first rationalize its prediction (\textit{explain then predict}) and then provide a label, versus rationalizing after making its label prediction (\textit{predict then explain}). Using GPT-4, for MultiRC, SciFact, and HealthFC, changing the request order caused a significant difference in the label prediction performance while Gemini-1.5 showed no significant differences. Using GPT-4, the predictive performance difference is mostly in favor of the \textit{predict then explain} condition, suggesting that in most cases constraining the model to pre-emptively justify its label has a slightly negative impact on its label accuracy. 
% Interestingly, the Gemini-1.5 model did not exhibit this pattern; in all cases except for WikiAttack, the differences between the two conditions were marginal. 

% Therefore, we observe mixed results across datasets from the effect of both forcing the model to provide evidence for its predictions, and providing human-annotated evidence to the model in the input prompt. However, comparing performance across these conditions allows us to begin to understand the underlying cause of model prediction and evidence retrieval errors.

\begin{table*}[]
\centering
\begin{tabular}{lcrrrr}
\hline
\multicolumn{2}{l}{\textbf{Explain then predict}} & \multicolumn{2}{c}{\textbf{Low evidence recall}}                                         & \multicolumn{2}{c}{\textbf{High evidence recall}}                                        \\ \hline
\multicolumn{2}{l}{\textbf{Evidence given}}       & \multicolumn{1}{c}{\textbf{Correct label}} & \multicolumn{1}{c}{\textbf{Incorrect label}} & \multicolumn{1}{c}{\textbf{Correct label}} & \multicolumn{1}{c}{\textbf{Incorrect label}} \\ \hline
\textbf{Dataset}                & \textbf{Model}              & \multicolumn{1}{c}{``Retrieval issue''}       & \multicolumn{1}{c}{``Unexplainable''}            & \multicolumn{1}{c}{``Confounding evidence''}         & \multicolumn{1}{c}{``Misinterpretation''}        \\ \hline
\multirow{2}{*}{MRC}         & GPT-4              & 0.12 (5)                                   & 0.24 (10)                                    & 0.22 (9)                                   & \textbf{0.41 (17)}                           \\
                             & Gemini-1.5         & 0.16 (6)                                   & 0.30 (11)                                    & 0.14 (5)                                   & \textbf{0.41 (15)}                           \\ \hline
\multirow{2}{*}{SF}          & GPT-4*             & 0.10 (5)                                   & 0.14 (7)                                     & \textbf{0.60 (30)}                         & 0.16 (8)                                     \\
                             & Gemini-1.5*        & 0.19 (9)                                   & 0.21 (10)                                    & \textbf{0.50 (24)}                         & 0.10 (5)                                     \\ \hline
\multirow{2}{*}{WA}          & GPT-4              & 0.00 (0)                                   & 0.06 (5)                                     & 0.34 (26)                                  & \textbf{0.60 (46)}                           \\
                             & Gemini-1.5         & 0.06 (5)                                   & 0.05 (4)                                     & \textbf{0.45 (40)}                         & 0.44 (39)                                    \\ \hline
\multirow{2}{*}{HFC}         & GPT-4*             & \textbf{0.73 (40)}                         & 0.07 (4)                                     & 0.13 (7)                                   & 0.07 (4)                                     \\
                             & Gemini-1.5         & \textbf{0.44 (24)}                         & 0.35 (19)                                    & 0.11 (6)                                   & 0.09 (5)                                     \\ \hline
\multirow{2}{*}{EI}          & GPT-4              & 0.26 (5)                                   & \textbf{0.32 (6)}                            & 0.16 (3)                                   & 0.26 (5)                                     \\
                             & Gemini-1.5*        & \textbf{0.39 (15)}                         & 0.16 (6)                                     & 0.16 (6)                                   & 0.29 (11)                                    \\ \hline
\end{tabular}
\caption{Contingency table of label prediction errors in \textit{explain then predict} condition; split by low vs. high evidence recall and correct vs. incorrect prediction in \textit{explanation given} condition. Asterisks denote significant differences via the Fisher exact test with p-value 0.05. An interpretation of each contingency is presented in quotes.}
\label{tab:label_error_reason_table}
\end{table*}

\subsection{Is evidence retrieval error correlated with label prediction error?}
\label{subsec:rationale_vs_label}

Table \ref{tab:correctness_table} shows the mean self-rationalization performance (compared to human-annotated evidence) for correct and incorrect label predictions made by each model, in the \textit{explain then predict} condition. Figure \ref{fig:label_rationale_whisker_plots} illustrates the distribution of rationale F1s under this condition. 

For GPT-4, there is a strong correlation between evidence F1 and label accuracy in SciFact, WikiAttack, and Evidence Inference, with respective evidence-F1 differences of 0.65, 0.70, and 0.24. It is likewise strong for Gemini-1.5 in SciFact and WikiAttack, with evidence-F1 differences of 0.66 for both. With respect to SciFact, hardly any incorrect predictions display any level of correct alignment. Again, this effect is consistent across both models.

This result suggests that for certain datasets (SciFact and WikiAttack), the primary challenge is in retrieving the correct evidence, and when this can be done the label can be predicted with high accuracy. There are others (MultiRC, Evidence Inference in the case of Gemini), where evidence retrieval is successful, and the challenge lies more in interpreting the correct answer from that evidence. HealthFC is an outlier, with both models performing poorly on evidence retrieval, for both correct and incorrect label predictions. 

%\subsubsection{Why does the model make prediction errors?}
\subsection{Why does the model make prediction errors?}
\label{subsec:why_prediction_errors}

When the model makes prediction errors, to what extent can we attribute these errors to missing evidence or misinterpretation of correct evidence? Is it more common for the model to miss key evidence entirely (which is difficult to correct via prompt engineering), or to misinterpret relevant evidence (which can be addressed by reinspection approaches)? Table \ref{tab:label_error_reason_table} explores these questions by breaking down label prediction errors made in the \textit{explain then predict} condition by (1) the evidence recall relative to human-annotated evidence in this condition, and (2) the model's label accuracy on those corresponding items on the \textit{evidence given} condition. This division lets us ask: \textbf{given that the model was wrong, did it identify the correct evidence, and would it still have been wrong if it had done so?}

In this division, the (low-recall, correct label) condition represents cases where the model's label prediction mistake would have been overturned by presenting it with the human rationale, meaning that we can attribute the mistake to the model having failed to retrieve the correct evidence. The (low-recall, incorrect label) condition represents cases where the model's mistake persists with or without the human rationale, making it impossible to attribute. The (high-recall, correct label) condition implies that there was extra confounding evidence the model picked up in addition to recovering the human rationale, which caused its label prediction error. Finally, the (high-recall, incorrect label) condition represents scenarios where the model successfully recovered the human rationale, but made the incorrect prediction regardless (misinterpretation). 

We find that in three of out five datasets (MultiRC, SciFact, and WikiAttack), label prediction errors are mostly associated with high rationale recall (misinterpretation and confounding evidence) rather than low recall (missing evidence). In MultiRC and WikiAttack, the model is most likely to recover the human-annotated evidence but be unable to interpret it properly, while for SciFact the model is likely to recover both the human rationale and extraneous evidence, which causes it to produce an incorrect prediction. For two datasets, HealthFC and Evidence Inference, a majority of prediction errors are associated with low evidence recall. In both models' cases, providing human evidence to the model would produce a correct label for HealthFC. For Evidence Inference, this remains true only for Gemini-1.5. This means that HealthFC is the only dataset for which we can attribute a majority of errors to missing key evidence that the model had the capacity to interpret. In other cases, the model is more likely to recover the relevant evidence and then misinterpret it.

\begin{table*}[]
\centering
\begin{tabular}{llrrrr}
\hline
\multicolumn{2}{l}{\textbf{Evidence occluded}}   & \multicolumn{2}{c}{\textbf{Label incorrect}}                                              & \multicolumn{2}{c}{\textbf{Label correct}}                                                \\ \hline
\multicolumn{2}{l}{\textbf{Evidence given}}      & \multicolumn{1}{c}{\textbf{Label correct}} & \multicolumn{1}{c}{\textbf{Label incorrect}} & \multicolumn{1}{c}{\textbf{Label correct}} & \multicolumn{1}{c}{\textbf{Label incorrect}} \\ \hline
\textbf{Note}        & \multicolumn{1}{c}{Model} & \multicolumn{1}{c}{``Retrieval issue''}          & \multicolumn{1}{c}{``Uninterpretable evidence''}          & \multicolumn{2}{c}{``Other viable evidence present''}                                         \\ \hline
\multirow{2}{*}{MRC} & GPT-4*                    & 0.22 (19)                                  & 0.14 (12)                                    & \textbf{0.60 (52)}                         & 0.05 (4)                                     \\
                     & Gemini-1.5                & 0.26 (29)                                  & 0.05 (6)                                     & \textbf{0.49 (55)}                         & 0.20 (22)                                    \\ \hline
\multirow{2}{*}{SF}  & GPT-4*                    & 0.37 (25)                                  & 0.10 (7)                                     & \textbf{0.51 (34)}                         & 0.01 (1)                                     \\
                     & Gemini-1.5*               & \textbf{0.42 (32)}                         & 0.09 (7)                                     & 0.41 (31)                                  & 0.08 (6)                                     \\ \hline
\multirow{2}{*}{WA}  & GPT-4                     & \textbf{0.47 (7)}                          & 0.20 (3)                                     & 0.13 (2)                                   & 0.20 (3)                                     \\
                     & Gemini-1.5                & \textbf{0.46 (13)}                         & 0.04 (1)                                     & 0.32 (9)                                   & 0.18 (5)                                     \\ \hline
\multirow{2}{*}{HFC} & GPT-4*                    & 0.30 (45)                                  & 0.01 (2)                                     & \textbf{0.65 (97)}                         & 0.04 (6)                                     \\
                     & Gemini-1.5*               & 0.21 (45)                                  & 0.06 (13)                                    & \textbf{0.63 (133)}                        & 0.09 (20)                                    \\ \hline
\multirow{2}{*}{EI}  & GPT-4*                    & 0.16 (8)                                   & 0.08 (4)                                     & \textbf{0.76 (38)}                         & 0.00 (0)                                     \\
                     & Gemini-1.5*               & 0.18 (11)                                  & 0.03 (2)                                     & \textbf{0.77 (48)}                         & 0.02 (1)                                     \\ \hline
\end{tabular}
\caption{Contingency table of low human rationale recall instances in \textit{explain then predict} condition; split by correct vs. incorrect label prediction in \textit{evidence occluded} and \textit{evidence given} conditions. Asterisks denote significant differences via the McNemar test with p-value 0.05. Intepretations provided in quotes.}
\label{tab:low_recall_reason_table}
\end{table*}

%\subsubsection{Why does the model miss key evidence?}
\subsection{Why does the model miss key evidence?}

Similar to Section \ref{subsec:why_prediction_errors}, we can use the results of different prompting paradigms to ask why the model misses key evidence. In particular, we can investigate the following hypothesis: \textbf{the model tends to miss evidence primarily when it is unable to interpret it correctly.}

Table \ref{tab:low_recall_reason_table} represents a contingency table of low-evidence-recall instances from the \textit{explain then predict} condition, divided by label prediction error in the \textit{evidence occluded} and \textit{evidence given} conditions. The (occluded incorrect, given correct) contingency represents cases where the human evidence was both necessary and sufficient for the model to correctly predict the label, while (occluded incorrect, given incorrect label) represents cases where the human evidence was not unnecessary, but also not sufficient to predict the label. Finally, either contingency where the label prediction was correct for the explanation-occluded condition is one where there was additional viable evidence beyond the human-annotated evidence, rendering it unnecessary, and low evidence recall less of a real error. Thus, the question being asked here is: \textbf{given that the model missed key evidence, was that evidence necessary and would it have been sufficient if it had been found?}

If misses of necessary human-annotated evidence were mostly associated with them being uninterpretable by the model, a majority of cases would fall in the (occluded incorrect, given incorrect) contingency. Instead, Table \ref{tab:low_recall_reason_table} shows that in a majority of cases, datasets fall into the (occluded correct, given correct) contingency, meaning that models mostly fail to retrieve human-annotated evidence when there is additional viable evidence that the model can use to successfully predict the label. 

Even when no such additional viable evidence is possible (incorrect label in the occluded condition), a majority of cases for all datasets are ones where the label is correct in the evidence-given condition, meaning the model does know how to correctly interpret the key evidence, it simply fails to retrieve it in the \textit{explain then predict} condition. In WikiAttack alone does this contingency represent a majority of overall cases. Hence, the hypothesis that retrieval failures are associated with evidence the model cannot interpret correctly is not supported in our analysis.

\section{Discussion}

Consider a human-model collaborative system such as an interpretable LM-driven moderation system, grounded in the mechanism of extractive self-rationalization. For verification to be possible in such a system, the LM must be able to consistently extract explanatory evidence from the input document, and that evidence should display external faithfulness, in that if it can be proven incorrect then the model's prediction should also be incorrect as well. As error modes go, it is better for the model to identify correct evidence and then misinterpret it than to miss evidence entirely, as this is more correctable via prompting or human inspection. And if the model is to miss evidence, it is better for it to be able to interpret it correctly if provided, as this could be supported by additional within-document retrieval support e.g. \cite{singhania_recall_2024}. The most irrecoverable outcome is the model missing evidence that it cannot interpret properly at all. 
% As error modes go, we prefer the more correctable retrieve-evidence-correctly-then-misinterpret mode over the miss-evidence-entirely mode. And if the model is going to miss evidence, we prefer for it to be able o 

The results shown above are largely supportive of these requirements. We find that models can mostly reliably quote evidence from the input, and that for at least some datasets, evidence retrieval performance is correlated with label prediction performance. We find that label errors are mostly associated with either confounding evidence or with missing evidence that could at least be interpreted correctly if it had been retrieved. Finally, when the model misses key evidence, we find it mostly associated with  the presence of other viable evidence, meaning that it is not truly missing evidence at all in most apparent cases. 

This is a hopeful outcome for any verification system based on self-rationalization. While the properties (and average document lengths) of the data in question has a major impact on the viability of this mechanism, there are at least some datasets for which it will tend to work well and potentially serve as the basis for such a system. Further work might develop an evaluation protocol to determine whether or not a given dataset falls into this category. 

\section{Conclusion}

In this work, we investigate the relationship between prediction and extractive rationalization in few-shot learning. We find that there is a strong correlation between the classification accuracy and agreement with human-annotated rationales. Furthermore, we find that significantly more model error is attributable to imprecise rationalization than incomplete rationalization, a positive sign for downstream methods based on this mechanism.

% References and End of Paper
% These lines must be placed at the end of your paper

\end{document}